\documentclass{article} % For LaTeX2e
\usepackage{iclr2027_conference,times}
\usepackage{graphicx}
\usepackage{multirow}
\usepackage[T1]{fontenc}
\usepackage{amsmath,amsfonts,bm}

\def\eqref#1{equation~\ref{#1}}
\def\1{\bm{1}}

\DeclareMathAlphabet{\mathsfit}{\encodingdefault}{\sfdefault}{m}{sl}
\SetMathAlphabet{\mathsfit}{bold}{\encodingdefault}{\sfdefault}{bx}{n}

\usepackage{hyperref}
\usepackage{url}
\usepackage{cleveref}

\renewcommand{\paragraph}[1]{\textbf{#1}}
\usepackage{enumitem}
\setlist[itemize]{noitemsep, topsep=0pt, partopsep=0pt, leftmargin=5mm}
\setlist[description]{noitemsep, topsep=4pt, partopsep=0pt,leftmargin=5mm}
\usepackage{caption}
\usepackage[most]{tcolorbox}
\tcbuselibrary{breakable,listings}

\newtcblisting{promptbox}[2][]{%
  enhanced,
  breakable,
  colback=gray!5,
  colframe=black!70,
  colbacktitle=gray!25,
  boxrule=0.6pt,
  arc=1.5mm,
  left=2mm,
  right=2mm,
  top=2mm,
  bottom=1.5mm,
  title={#2},
  fonttitle=\bfseries,
  coltitle=black,
  boxed title style={
    colback=gray!25,
    colframe=black!50,
    boxrule=0.5pt,
    arc=1mm
  },
  listing only,
  verbatim,
  #1
}

\title{Medical Image Alignment Assessment as a Test of Generalist Visual Reasoning in Frontier Multimodal Models}

\author{Ross Callaghan\textsuperscript{1,2}, Niannu Gao\textsuperscript{2}, Hojjat Azadbakht\textsuperscript{2} \& Hui Zhang\textsuperscript{1} \\
\textsuperscript{1}Hawkes Institute, University College London, London, UK\\
\textsuperscript{2}AINOSTICS Ltd., Manchester, UK\\
}

\iclrfinalcopy % Uncomment for camera-ready version, but NOT for submission.
\begin{document}

\maketitle

\begin{abstract}
    Frontier multimodal large language models (MLLMs) are increasingly positioned as general purpose visual reasoners as part of the quest for artificial general intelligence. 
    A key test of this generality is whether they can perform novel visual judgments that humans can make reliably from visual evidence and task instructions, without task-specific parameter optimisation. 
    We investigate this question through the task of medical image alignment assessment, where the goal is to establish whether there is anatomical correspondence between two images. 
    Assessing spatial alignment is a fundamental component of medical imaging pipelines across within-subject, between-subject, and cross-scanner settings.
    Human visual assessment of image alignment is still the gold standard and most common approach; however, it requires trained operators and is impractical to scale for large datasets. Existing automated methods using conventional image quality metrics or convolutional neural networks (CNNs) are generally designed for particular modalities, anatomical regions, or alignment tasks, not generalisable readily to new settings.
    We evaluate recent generations of MLLMs on two exemplar medical image alignment tasks, varying both prompting strategies and image-presentation methods.
    We compare against a smaller, locally fine-tuned MLLM and a task-specific CNN, allowing us to examine the trade-off between frontier general purpose models that cannot be used locally and smaller models that require specific task optimisation but can be used locally. 
    Our results show that we are reaching an inflection point, where frontier MLLMs can now effectively perform visual assessment of medical image alignment.
    Models released only a few months ago generalise poorly and, in some settings, perform barely above chance, whereas GPT-6 achieves over 85\% across almost all scenarios tested, including under zero-shot evaluation.
    Fine-tuned local models can match or exceed frontier-model performance on the tasks on which they are trained, but transfer substantially less effectively to unseen settings.
    These findings identify medical image alignment as a useful test bed for generalist visual reasoning. They also suggest that frontier multimodal models are beginning to acquire transferable visual assessment capabilities that could support a common quality-control mechanism across heterogeneous medical-imaging pipelines.
\end{abstract}

\section{Introduction}
\label{sec:intro}
Generalist visual reasoning is a key aspect of the push for artificial general intelligence in multimodal large language models (MLLMs). 
More precisely, this means that MLLMs should be able to perform novel visual tasks that humans can do reliably from visual evidence and task instructions, without task-specific parameter optimisation. 
We investigate whether this is the case for frontier MLLMs through the task of medical image alignment assessment. 

Medical imaging is critically important in modern healthcare, helping to guide diagnostic, prognostic and treatment decisions. Establishment of spatial alignment between medical images is a crucial part of many image analysis pipelines, ensuring anatomical correspondence between images across within-subject, between-subject and cross-scanner settings. This correspondence is key to almost all aspects of medical image analysis; for instance, we need to check raw data for motion between volumes \citep{andersson2016, legouhy2026, friston1996, picard1997, ye2014} and ensure that images are aligned for multimodal fusion \citep{greve2014, greve2016, dore2019, kim2024} and longitudinal analysis \citep{reuter2012, fox1997,bosc2003}.  

%In the within-subject setting, image alignment assessment is important in multi-frame acquisitions to detect patient motion between frames \citep{andersson2016, legouhy2026, friston1996, picard1997, ye2014}. For multimodal data, image alignment is necessary for fusion of data across modalities, for instance ensuring positron emission tomography (PET) and magnetic resonance imaging (MRI) images are in the same space to link the anatomical information in the MRI to the physiological information in the PET \cite{greve2014, greve2016, dore2019, kim2024}.
%Furthermore registration enables within-subject longitudinal analysis \cite{reuter2012, fox1997,bosc2003} and can be used to correct for intra-operative brain shift \cite{clatz2005, bayer2017}. 

%Between-subjects, images are aligned to bring subjects into a common reference space such as the MNI-ICBM152 space \cite{manera2020, fonov2011a}. Bringing subjects into a shared space further enables atlas-based segmentation, propagating a segmentation defined in the atlas space back to the subject space \cite{wang2013, aljabar2009, heckemann2006} and  analysis of differences between-subjects, through approaches such as voxel-based morphometry \cite{douaud2007, good2001}. 

Since image alignment is a cornerstone of so many image analysis pipelines, it is critical to assess alignment and ensure it is sufficient to ensure the accuracy of downstream processing. Indeed, \cite{dadar2018} found typical failure rates of around 5-10\% when investigating methods for alignment of T1-weighted (T1w) MRI, a type of MRI that is used to visualise anatomical details, to a template space. 
While image alignment assessment is essential to ensure that any alignment errors do not propagate to downstream analyses, it is inherently difficult, particularly in the multimodal setting \citep{rohlfing2012, bierbrier2022}. 

%There is an important distinction to make in relation to registration QC between the within-subject and between-subject settings. Within-subject, since the anatomy should be consistent between the images, there exists a ``correct'' registration to bring images into anatomical correspondence and the task of registration QC consists of assessing the accuracy of this transform. Between-subject, there are fundamental differences in the underlying anatomy and as such establishing direct anatomical correspondence is not well posed. In this case, the task of registration QC becomes rather establishing the plausibility of the registration rather than a strict accuracy. 

\begin{figure}
    \centering
    \includegraphics[width=\linewidth]{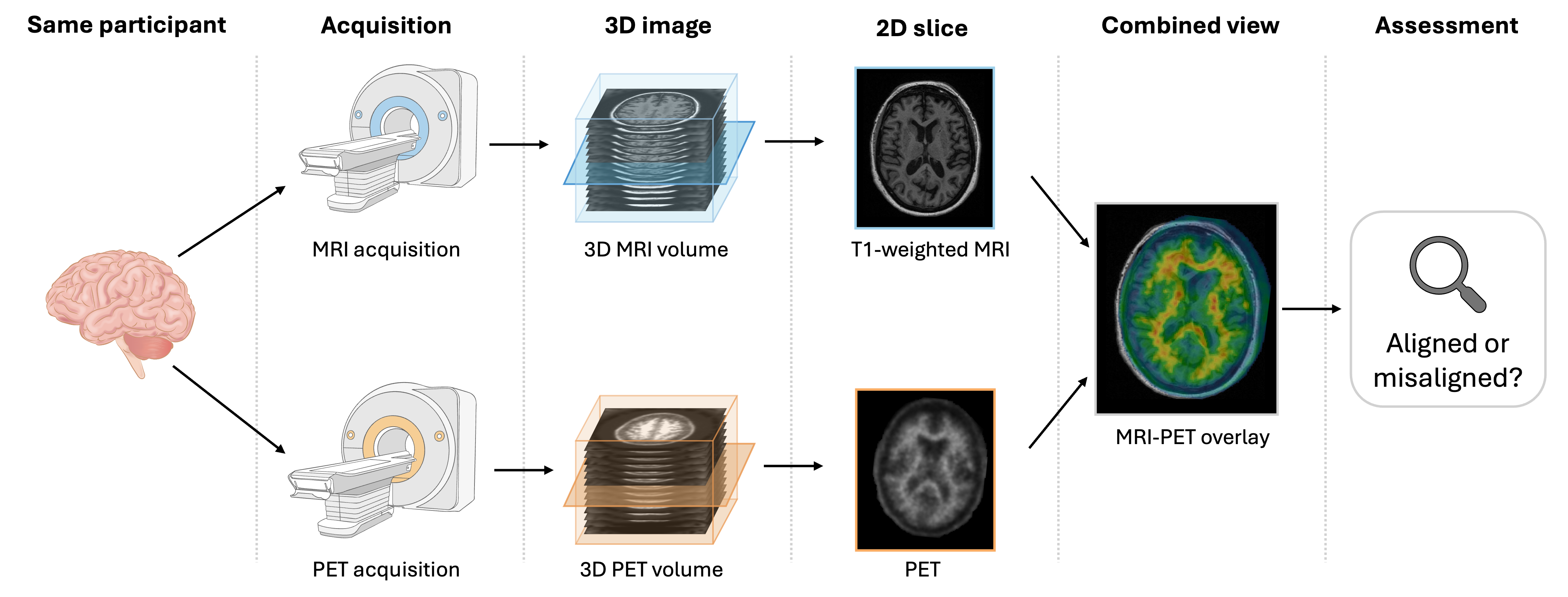}
    \caption{Illustration of the medical image alignment assessment problem.}
    \label{fig:image_alignment_example}
\end{figure}

In practice, the gold standard for image alignment assessment is manual inspection. Typically this will be achieved by overlaying the two images using an image viewing tool to visually assess the accuracy or plausibility of the alignment. This can include directly overlaying the images with translucency and overlaying anatomical cues, such as segmentation outlines from one image onto another. \Cref{fig:image_alignment_example} shows an example of this process.  Visual assessment, however, requires trained assessors and is time consuming. This makes it impractical for large datasets.

Efforts have been made to automate the image alignment assessment process, attempting to overcome many of the limitations of manual visual assessment; however, automating this process is complex. Some approaches use traditional image quality or segmentation overlap metrics \citep{rohlfing2012, bosma2023, esteban2017}, while deep-learning approaches have recently been proposed for image alignment assessment for quality control of registration \citep{denisdesenneville2020, fonov2022, lallement2023, salari2023}, a task aiming to place a pair of images into alignment. These methods, however, are generally designed for particular modalities, anatomical regions, or registration tasks and do not generalise readily to new settings.

%Recently, deep-learning approaches have been proposed to estimate registration QC metrics directly from images. RegQCNET uses a 3D convolutional neural network (CNN) to estimate the quality of an affine alignment between a T1w MRI of the brain and a template, estimating a root mean square error (RMSE) between the registered MRI and the template \cite{denisdesenneville2020}. Similarly, DARQ uses a 2D CNN to output a binary pass-fail decision on affine registration quality between T1w brain MRI and a template space \cite{fonov2022}. Outside of brain MRI, Lallement et al. 2022 \cite{lallement2023}, use CNNs to detect errors in deformable image registration between longitudinal head and neck CT in radiotherapy. Salari et al., 2023 \cite{salari2023} train a 3D Swin-UNETR-based model to predict registration error maps between MRI and intra-operative ultrasound. These approaches are generally developed for a particular modality pair, anatomy and transformation class, and require task-specific training using expert QC labels, synthetic perturbations or registration-error targets.

The task of MRI-PET alignment exemplifies the challenges inherent in developing automated image alignment assessment approaches. 
MRI and PET depict different physical and biological properties, and therefore lack a simple relationship between corresponding voxel intensities. 
Moreover, this relationship may differ across radioactive chemical tracers used in a PET exam and disease states, for instance different contrast patterns of amyloid PET (a type of PET exams using tracers that preferentially bind to amyloid plaques, the hallmark of Alzheimer's) between Alzheimer's disease patients and controls. 
Segmentation or landmark based approaches are equally limited in this context because the comparatively low spatial resolution and tracer-dependent contrast of PET make these features difficult to obtain reliably in both modalities. 
Importantly, these limitations do not prevent visual assessment of MRI-PET alignment. Human readers can often fairly easily detect anatomical misalignment by checking the correspondence of anatomical cues derived from the structural MRI with the PET image, such as in \Cref{fig:image_alignment_example}. 

%In this case, the registration algorithm optimised normalised mutual information (NMI), showing that optimal image similarity does not necessarily lead to anatomical correspondence and that image alignment quality assessment may be viewed not only as a problem of measuring image similarity, but also as one of interpreting whether the anatomical relationships depicted across two images are spatially plausible.

\begin{figure}[ht]
\centering
\includegraphics[width=\linewidth]{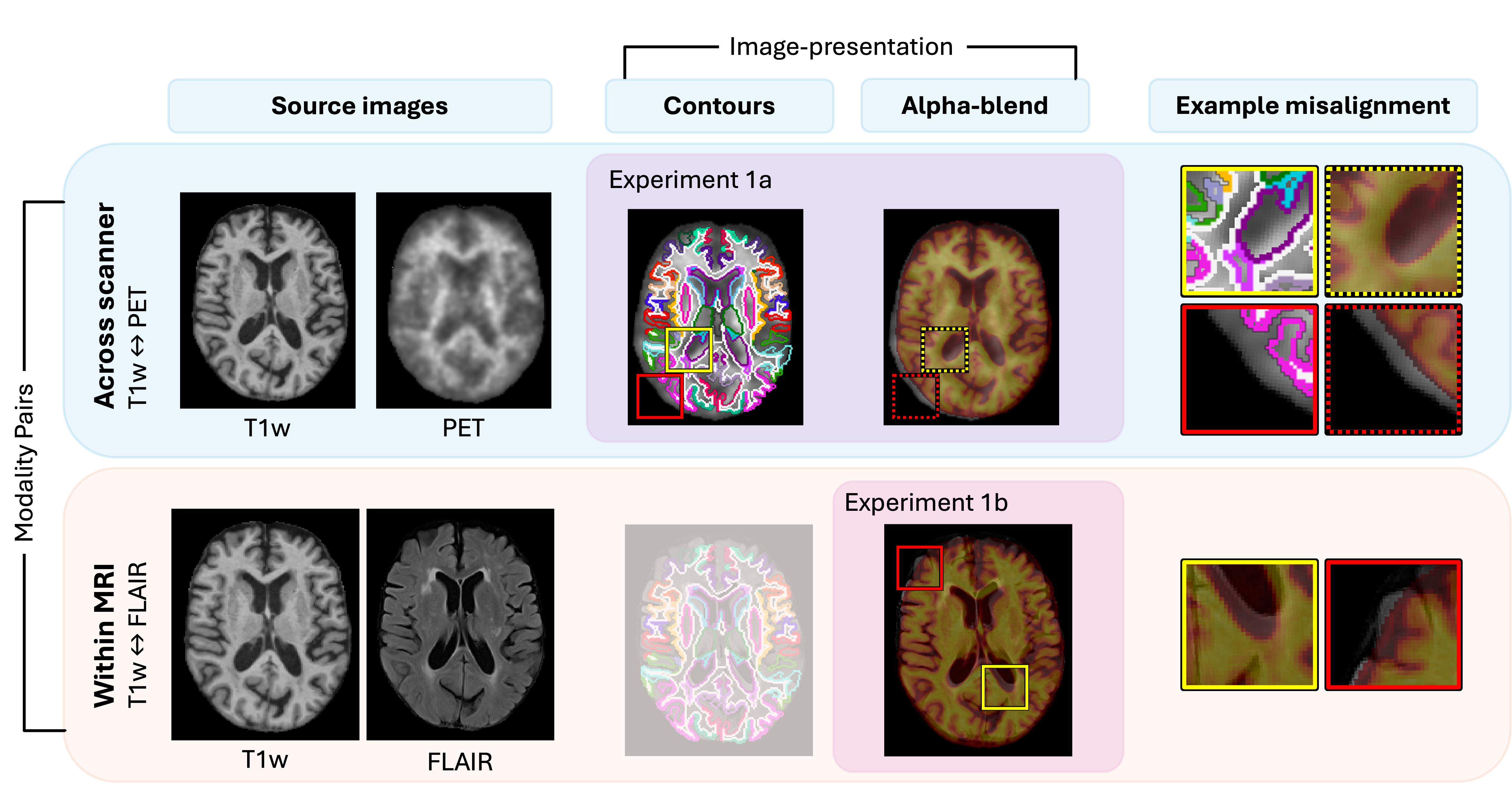}
\caption{Overview of the medical image alignment assessment method. We test image alignment assessment on two modality pairs: cross-scanner T1w MRI-PET and within MRI T1w-FLAIR alignment. 
%In Experiment 1 we evaluate a range of models, prompting and image presentation techniques on the MRI-PET task. In Experiment 2 we test how well the best performing of these generalise to T1w-FLAIR.  
We highlight how different modality pairs and image-presentation techniques can affect the alignment assessment task. Yellow zoom boxes show misalignment of ventricles, something which is easy to spot for MRI-PET contours and T1w-FLAIR alpha-blend (solid yellow boxes) but harder to see in MRI-PET alpha-blend (dashed yellow box). Red zoom boxes show brain-edge misalignment which is easy to spot in MRI-PET alpha-blend (dashed red box) but harder to definitively see in MRI-PET contours (solid red box). 
}
\label{fig:idea_example}
\end{figure}

Recently, multimodal large-language models (MLLMs) have demonstrated increasingly strong generalised image understanding \citep{yue2023,yue2024,geminiteam2023,openai2024}, including one recent study utilising MLLMs for visual quality control in a wider agentic framework \citep{han2026}. 
We test how well this generalist visual reasoning can be used to replicate a human-like image alignment assessment process in two exemplar medical imaging settings as outlined in \Cref{fig:idea_example}. 
We further examine the tradeoff between frontier API-based general purpose models and local models that require specific task optimisation by comparing frontier MLLMs to a smaller, locally fine-tuned MLLM and a task-specific CNN.

\section{Experimental design}
\label{sec:methods}
%We test a range of recent frontier MLLMs for misalignment detection in three experiments. The primary experiments test MLLMs across two medical image alignment settings, as outlined in \Cref{fig:idea_example}, while a secondary experiment compares frontier MLLMs against local models:
To test the generalist visual reasoning of frontier MLLMs, our first experiment involves testing a range of models in two medical image alignment settings to establish how well they can perform this task. We then investigate whether frontier MLLMs can now outperform task-specific locally trained models in our second experiment: 
% \begin{itemize}
% \item \textbf{Experiment 1: MLLMs as medical image alignment assessors} 
% \begin{itemize}
%     \item \textbf{Experiment 1a: MRI-PET alignment}, a challenging setting where although the two images come from the same subject, they convey different information. We use this as a test bed to probe the impact of various prompting/input choices on model performance.
%     \item \textbf{Experiment 1b: T1w-FLAIR (fluid attenuated inversion recovery) MRI alignment}, a theoretically simpler task where the two images both show aspects of brain structure. We test how well best performing settings from the MRI-PET experiment transfer to this dataset.
% \end{itemize}
%     \item \textbf{Experiment 2: Comparison to local models} we test a smaller locally fine-tuned MLLM and a task-specific CNN, to establish whether MLLMs can outperform task optimised models. 
% \end{itemize} 
\begin{description}[itemsep=0em]
\item[Experiment 1] Testing MLLMs as medical image alignment assessors
\begin{description}[itemsep=0em]
    \item[Experiment 1a: MRI-PET alignment] a challenging setting where although the two images come from the same subject, they convey different information. We use this as a test bed to probe the impact of various prompting/input choices on model performance.
    \item[Experiment 1b: T1w-FLAIR (fluid attenuated inversion recovery) MRI alignment] a theoretically simpler task where the two images both show aspects of brain structure. We test how well best performing settings from the MRI-PET experiment transfer to this dataset.
    \item[Experiment 1c: Real Misalignment] Experiments 1a and 1b use synthetic examples in which we introduce misalignment to well-aligned data. In this experiment we test MLLMs' ability to detect misalignment in cases of actual alignment method failure. 
\end{description}
\item[Experiment 2] Comparison to local models \\We test a smaller locally fine-tuned MLLM and a task-specific CNN to establish whether, and in which scenarios, API-based frontier MLLMs outperform local task-specific models.

\end{description}

% \subsection{Experiment 1: MLLMs as medical image alignment assessors}
% \label{sec:exp1_method}

\subsection{Experiment 1a: MRI-PET alignment}
\label{sec:exp1a_method}
To benchmark the performance of MLLMs for image alignment assessment, we use the task of detecting misalignment between T1w MRI and PET pairs, a task which is relatively easy for human visual assessors but difficult to robustly implement with traditional metrics. We start from well-aligned MRI-PET pairs that have been manually assessed, introduce synthetic misalignments through random rigid transformations and test the MLLMs' ability to detect this misalignment. 

There are a number of axes of variation that can be explored when it comes to utilising MLLMs for image alignment. Here, we explore three: MLLM model choice and their performance over time, zero-shot vs. few-shot prompting, and image presentation method. 

\paragraph{MLLM model choice}
% \label{sec:model_choice}
We test a number of frontier MLLM models released over recent months and years as detailed in \Cref{tab:models}. We additionally test a local MLLM that is trained for performance on medical text and image comprehension, MedGemma \citep{sellergren2026}.

\begin{table}[ht]
    \centering
    \begin{tabular}{ccc}
        Model Family & Model & Release Date \\
        \hline
        \multirow{6}{4em}{GPT} & GPT-5.2 & Dec `25 \\
        & GPT-5.4 & Mar `26 \\
        & GPT-5.6 Luna & Jul `26 \\
        & GPT-5.6 Terra & Jul `26 \\
        & GPT-5.6 Sol & Jul `26 \\
        & GPT-6 Astra & Sep `26 \\\hline
        \multirow{5}{4em}{Gemini} & Gemini 3 Flash & Dec `25 \\
        & Gemini 3.1 Pro & Feb `26 \\
        & Gemini 3.6 Flash & July `26 \\
        & Gemini 3.7 Flash & Aug `26 \\
        & Gemini 3.8 Flash & Sep `26 \\\hline
        \multirow{4}{4em}{Claude} & Claude Opus 4.6 & May `26 \\
        & Claude Sonnet 5 & Jun `26 \\
        & Claude Opus 5 & Jul `26 \\
        & Claude Fable 5.1 & Sep `26 \\\hline
        \multirow{1}{4em}{MedGemma\textsuperscript{1}} & MedGemma 1.5 4B & May `26 \\\hline
        \multirow{1}{4em}{ResNet\textsuperscript{2}} & ResNet-18 11M & `16
    \end{tabular}
    \caption{Main models tested in this work. \textsuperscript{1} MedGemma tested both as zero/few-shot MLLM in Experiment 1 and locally fintuned model in Experiment 2. \textsuperscript{2} ResNet tested as locally finetuned model in Experiment 2.  }
    \label{tab:models}
\end{table}

\begin{figure}
    \centering
    \includegraphics[width=0.65\linewidth]{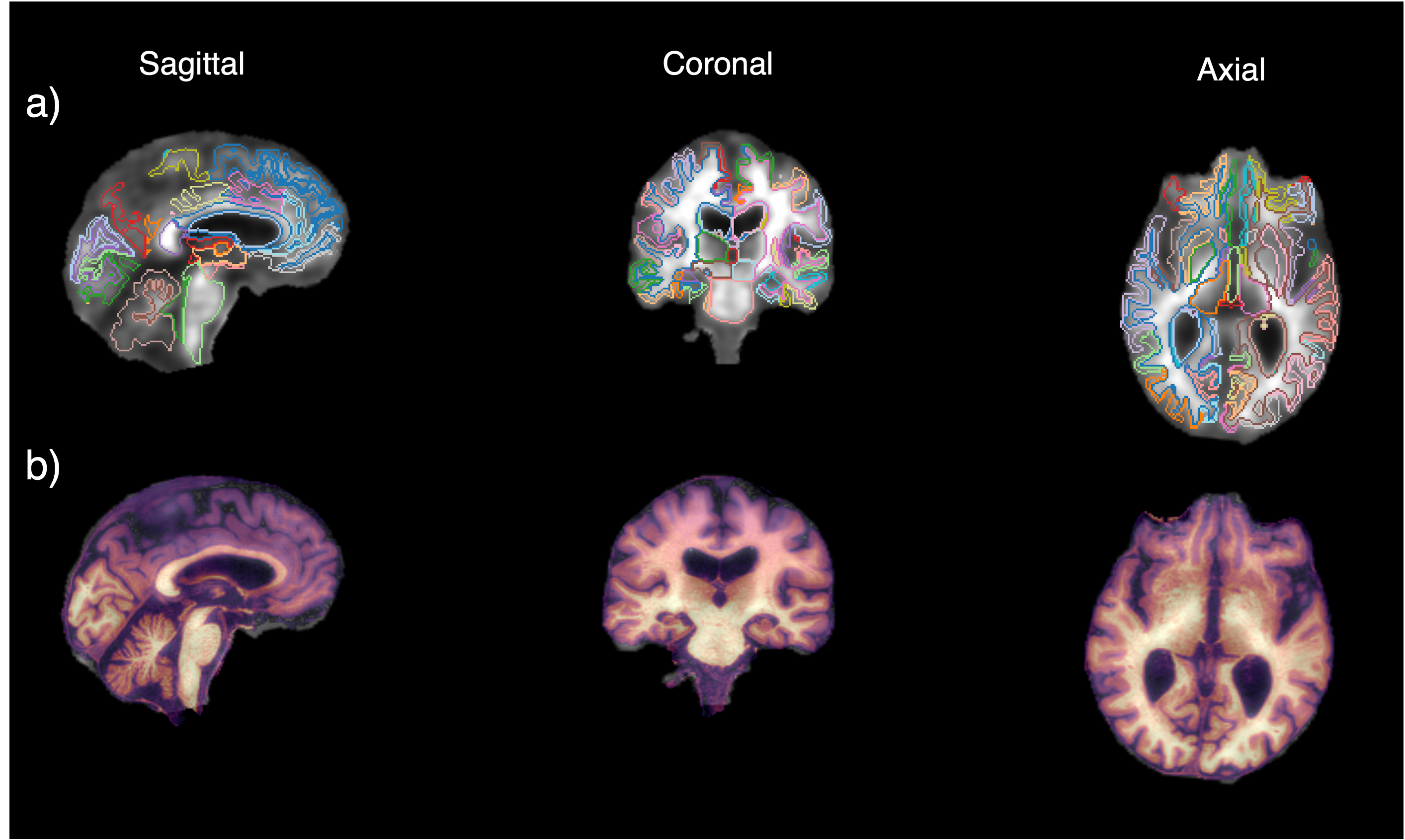}
    \caption{Example input images in an aligned case for a) contours presentation with anatomical segmentation contours from T1w MRI overlaid on the PET and b) alpha-blend presentation with the T1w image overlaid directly on the pet with 50\% opacity.}
    \label{fig:input_examples}
\end{figure}

\paragraph{Zero-shot vs.\ few-shot prompting}
\label{sec:zero_vs_few}
We test the zero-shot performance of MLLMs by simply prompting them with a question to determine whether the images are aligned, followed by the actual test image (see \Cref{fig:input_examples} for example input images).
Additionally, we test few-shot performance using an in-context learning approach similar to \cite{han2026} where we embed a few example images with labels into the prompt. Example images come from an additional dataset including subjects not used for the other experiments. 
See Appendix \ref{sec:appendix_prompts} for examples of the prompts used.

\paragraph{Image presentation method}
\label{sec:image_pres_method}
We test two different methods of presenting the images to the MLLMs inspired by how humans might visually assess alignment to see how it affects performance. 
Firstly, simply overlaying the T1w MRI on the PET with partial transparency. 
We term this approach "alpha-blend". Secondly, overlaying outlines of anatomical segmentation on the T1w MRI on top of the PET. We term this approach "contours". 
In each case, we present the models with three views of the brain, a central slice from each of the axial, sagittal and coronal views. 
\Cref{fig:input_examples} demonstrates these two approaches. 

% \begin{figure}[ht]
% \centering
% \includegraphics[width=0.6\linewidth]{images/qc_examples_2.png}
% \caption{Example image presentation methods. a) T1w MRI overlaid on PET with partial transparency. b) Anatomical segmentation from T1w MRI overlaid on PET }
% \label{fig:example_presentation}
% \end{figure}

\subsubsection{Data}
\label{sec:exp1a_data}
For experiments, we use data obtained from the Alzheimer’s Disease Neuroimaging Initiative (ADNI) database (adni.loni.usc.edu). 
%The ADNI was launched in 2003 as a public-private partnership, led by Principal Investigator Michael W. Weiner, MD. The primary goal of ADNI has been to test whether MRI, PET, other biological markers, and clinical and neuropsychological assessment can be combined to measure the progression of mild cognitive impairment (MCI) and early Alzheimer’s disease (AD). 
For the main MRI-PET test set, we use $^{18}$F-AV-45 (florbetapir) PET \citep{camus2012} and T1w MRI acquired from at the same session for 52 sessions, from 29 subjects from the ADNI database. 
%The PET images acquired were the "Co-reg, Avg, Std Img and Vox Siz, Uniform Resolution" preprocessed data which have undergone head motion correction, averaging, reorientaiton to a standard AC-PC alignment, SUVr normalisation and smoothing to a common resolution using scanner-specific 3D-Gaussian filters through ADNI's internal pipeline. 
The PET images acquired had undergone the standard ADNI preprocessing. 
We obtain the raw T1w MRI scan acquired at the same session for each PET image. 

\subsubsection{Ground truth alignment}
\label{sec:exp1a_ground_truth_alignment}
We align each subject's PET image to the corresponding T1w image using an in-house pipeline that runs three different rigid alignment algorithms and attempts to choose the best aligned method. 
The attempted alignment algorithms are: ANTs \citep{avants2008,avants2011}, \texttt{mri\_coreg} from FreeSurfer \citep{fischl2012} and Polaffini \citep{legouhy2026a}. 
ANTs and FreeSurfer are image-similarity based, using mutual information to optimise alignment, while Polaffini is segmentation-based, using SynthSeg \citep{billot2023, billot2023a} to segment both the T1w and PET.  
To select the best aligned method, we use mutual information multiplied by the dice similarity coefficient between SynthStrip \citep{hoopes2022a} brain masks derived from both the MRI and PET. 
As discussed above, image-similarity and segmentation-based metrics are not entirely reliable measures of image alignment, particularly in the context of MRI-PET. SynthSeg and SynthStrip, although developed to be robust, are not completely reliable in PET imaging, while mutual information is only a surrogate measure of image alignment. 
We manually visually inspect each output to ensure alignment before moving on with the experiments. 

\subsubsection{Introduced misalignment}
\label{sec:exp1a_introduced_misalignment}
For each manually verified aligned pair, we introduce synthetic misalignment by introducing a random rigid transformation to the PET image. We sample random translations in x,y,z $\in [2.5, 10]$ mm and rotations about x,y,z $\in [3.5,10]^{\circ}$, i.e. implicitly setting a misalignment classification threshold at 2.5mm, 3.5$^{\circ}$. For each pair, we randomly select whether to introduce a translation-only, rotation-only or combined misalignment and then further randomly sample along which axes the transform(s) should apply. 

\subsubsection{Performance evaluation}
\label{sec:exp1a_performance_evaluation}
The primary metric that we use to evaluate performance is accuracy of misalignment prediction from each MLLM against the ground truth label. For each MRI-PET pair, each MLLM is tested on the ground-truth aligned case and one random misaligned case, meaning that we have 104 total cases, 52 aligned and 52 misaligned. 

\subsection{Experiment 1b: T1w-FLAIR MRI Alignment}
\label{sec:exp1b_t1_flair_methods}
After benchmarking the MLLMs on MRI-PET alignment, we test a subset of MLLMs in the T1w-FLAIR setting. 
For this experiment, we simplify our model selection by taking two models per model family: one that is positioned as the 'frontier' model for that family, and one that is positioned as a 'cost-effective' model for that family. 
Furthermore, based on the results of Experiment 1a, we limit this evaluation to the best performing settings: alpha-blend image presentation with either zero-shot or one aligned and one misaligned example few-shot prompting. 
The same approach to introducing misalignment and performance evaluation is used as in \Cref{sec:exp1a_method}.

\subsubsection{Data}
\label{sec:exp1b_data}
To test the robustness of methods across different scanners and MRI acquisition parameters, we use data from three studies to test T1w-FLAIR alignment assessment. 
We use FLAIR and T1w MRI acquired at the same session for 11 sessions from 10 subjects from the ADNI database, as well as 13 sessions from 13 subjects from the A4 study \citep{sperling2023} and  16 sessions from 16 subjects from the MICCAI 2016 multiple sclerosis lesion segmentation challenge \citep{commowick2021}.

\subsubsection{Ground Truth Alignment}
\label{sec:exp1b_ground_truth_alignment}
We follow a similar approach as in \Cref{sec:exp1a_ground_truth_alignment}, however since T1w-FLAIR registration is generally more reliable, we use a single alignment pipeline based on ANTs to align skull-stripped images using SynthStrip.  

\subsection{Experiment 1c: Real Misalignment}
\label{sec:exp1c_real_misalignment}
To validate that the results observed in our synthetic misalignment experiment reflect the ability to detect the kind of misalignment seen in practice, we test MLLMs ability to detect misalignment in 30 subjects (from processing of the training set \Cref{sec:exp2_data}) where one of the alignment methods described in \Cref{sec:exp1a_ground_truth_alignment} failed human visual assessment. To distinguish genuine detection from always predicting misaligned, we include additional examples from 10 of these subjects using one of the other alignment methods in which alignment was successful. 

\subsection{Experiment 2: MLLM comparison against local models}
\label{sec:local_models}
To assess the benefits of supposed generalist MLLMs over task-specific models, we finetune a local MLLM, MedGemma, on the specific task of MRI-PET misalignment detection. 
Additionally, we finetune an ImageNet \cite{deng2009} pretrained ResNet-18, following the approach of DARQ \citep{fonov2022} as a standard CNN baseline. 
Each local model is trained on the MRI-PET task and with contour based image presentation, since this is the most challenging setting identified in Experiment 1a. 
We test how much data is required to achieve optimal performance by varying the training dataset size from 1\%-100\% of the training dataset outlined in \Cref{sec:exp2_data}.
We then test generalisability at each training set size on T1w-FLAIR task with contour based image presentation to test generalisability across modalities and the MRI-PET task with alpha-blend image presentation to test generalisability across image presentation methods. 
The same approach to introducing misalignment and performance evaluation is used as in \Cref{sec:exp1a_method}.

\subsubsection{Data}
\label{sec:exp2_data}
For training the local models, we use \textsuperscript{18}F-AV-45 PET and T1w MRI from ADNI the same as described in \Cref{sec:exp1a_data}. We use 2143 sessions from 1006 subjects as the total training set and 518 sessions from 125 subjects for validation. All subjects used for training and validation are distinct from the 29 used for testing. 
The same approach for ground-truth alignment is used as in \Cref{sec:exp1a_ground_truth_alignment}.

\section{Results}
\label{sec:results}
\subsection{Experiment 1a: MRI-PET alignment}
\label{sec:exp_1a_res}
Performance varies greatly across both models, zero/few shot promtping and image presentation method as detailed in \Cref{tab:results}. Across the board, GPT-6 Astra performs strongly, while Claude Fable 5.1 is best or second best in four of the six scenarios. 
\newcommand{\result}[3]{%
  \shortstack[c]{#1\\[-1pt]{\scriptsize #2/#3}}%
}
\begin{table}[ht]
    \centering
    \begin{tabular}{c|cccccc}
    \multicolumn{1}{r|}{Image Presentation} & \multicolumn{3}{c}{Contours} & \multicolumn{3}{c}{Alpha-blend} \\
    \multicolumn{1}{r|}{No.\ Shot Examples (a/m)} & 0/0 & 1/1 & 1/2 & 0/0 & 1/1 & 1/2 \\
    Model & & & & & & \\
    \hline
    GPT-5.2 & \result{57.7}{38.5}{76.9} & \result{63.5}{59.6}{67.3} & \result{58.7}{32.7}{84.6} & \result{67.3}{36.5}{98.1} & \result{88.5}{90.4}{86.5} & \result{\underline{91.3}}{86.5}{96.2} \\

    GPT-5.4 & \result{52.9}{13.5}{92.3} & \result{59.6}{28.8}{90.4} & \result{64.4}{36.5}{92.3} & \result{54.8}{9.6}{100.0} & \result{\underline{92.3}}{92.3}{92.3} & \result{86.5}{78.8}{94.2} \\

    GPT-5.6 Luna & \result{51.0}{1.9}{100.0} & \result{48.1}{21.2}{75.0} & \result{50.0}{7.7}{92.3} & \result{57.7}{15.4}{100.0} & \result{81.7}{98.1}{65.4} & \result{79.8}{98.1}{61.5} \\

    GPT-5.6 Terra & \result{51.0}{1.9}{100.0} & \result{50.0}{7.7}{92.3} & \result{53.8}{11.5}{96.2} & \result{71.2}{44.2}{98.1} & \result{83.7}{94.2}{73.1} & \result{86.5}{94.2}{78.8} \\

    GPT-5.6 Sol & \result{50.0}{0.0}{100.0} & \result{54.8}{13.5}{96.2} & \result{54.8}{15.4}{94.2} & \result{82.7}{69.2}{96.2} & \result{82.7}{98.1}{67.3} & \result{83.7}{98.1}{69.2} \\

    GPT-6 Astra & \result{\underline{74.0}}{48.1}{100.0} & \result{\textbf{87.5}}{75.0}{100.0} & \result{\textbf{85.6}}{71.2}{100.0} & \result{\textbf{87.5}}{75.0}{100.0} & \result{\textbf{97.1}}{94.2}{100.0} & \result{\textbf{97.1}}{94.2}{100.0} \\
    \hline

    Gemini 3 Flash & \result{59.0}{26.0}{92.0} & \result{71.2}{69.2}{73.1} & \result{59.6}{30.8}{88.5} & \result{83.7}{88.5}{78.8} & \result{60.6}{98.1}{23.1} & \result{78.8}{100.0}{57.7} \\

    Gemini 3.1 Pro & \result{62.5}{86.5}{38.5} & \result{66.3}{86.5}{46.2} & \result{58.7}{88.5}{28.8} & \result{\underline{85.6}}{92.3}{78.8} & \result{83.7}{98.1}{69.2} & \result{83.7}{98.1}{69.2} \\

    Gemini 3.6 Flash & \result{71.2}{67.3}{75.0} & \result{76.9}{67.3}{86.5} & \result{72.1}{51.9}{92.3} & \result{71.2}{100.0}{42.3} & \result{60.6}{100.0}{21.2} & \result{66.3}{100.0}{32.7} \\

    Gemini 3.7 Flash & \result{69.2}{57.7}{80.8} & \result{74.0}{59.6}{88.5} & \result{71.2}{86.5}{55.8} & \result{76.0}{98.1}{53.8} & \result{77.9}{100.0}{55.8} & \result{77.9}{98.1}{57.7} \\

    Gemini 3.8 Flash & \result{68.3}{44.2}{92.3} & \result{74.0}{55.8}{92.3} & \result{67.3}{84.6}{50.0} & \result{70.2}{96.2}{44.2} & \result{81.7}{98.1}{65.4} & \result{77.9}{98.1}{57.7} \\
    \hline

    Claude Opus 4.8 & \result{55.8}{30.8}{80.8} & \result{66.3}{44.2}{88.5} & \result{72.1}{67.3}{76.9} & \result{59.6}{19.2}{100.0} & \result{77.0}{60.0}{94.0} & \result{70.2}{44.2}{96.2} \\

    Claude Sonnet 5 & \result{52.9}{23.1}{82.7} & \result{56.7}{26.9}{86.5} & \result{58.7}{38.5}{78.8} & \result{51.0}{17.3}{84.6} & \result{72.1}{63.5}{80.8} & \result{74.0}{57.7}{90.4} \\

    Claude Opus 5 & \result{58.7}{86.5}{30.8} & \result{73.1}{67.3}{78.8} & \result{74.0}{71.2}{76.9} & \result{75.0}{80.8}{69.2} & \result{85.6}{92.3}{78.8} & \result{88.5}{92.3}{84.6} \\

    Claude Fable 5.1 & \result{\textbf{75.0}}{78.8}{71.2} & \result{\underline{79.8}}{69.2}{90.4} & \result{\underline{78.8}}{69.2}{88.5} & \result{72.1}{57.7}{86.5} & \result{89.0}{84.0}{94.0} & \result{\underline{91.3}}{84.6}{98.1} \\
    \hline

    MedGemma 1.5 4B & \result{50.0}{0.0}{100.0} & \result{53.8}{80.8}{26.9} & \result{51.9}{23.1}{80.8} & \result{50.0}{0.0}{100.0} & \result{52.9}{30.8}{75.0} & \result{48.1}{0.0}{96.2} \\
    \hline

\end{tabular}

    \caption{Misalignment detection accuracy across MLLM models, zero/few shot prompting and image presentation method. a/m refers to number of aligned/misaligned examples used in the few-shot case. Values reported are accuracy as a percentage with sensitivity/specificity below. Boldface numbers denote the best accuracy model per-column, underlined numbers the second-best.}
    \label{tab:results}
\end{table}

\begin{figure}[ht]
    \centering
    \includegraphics[width=\linewidth]{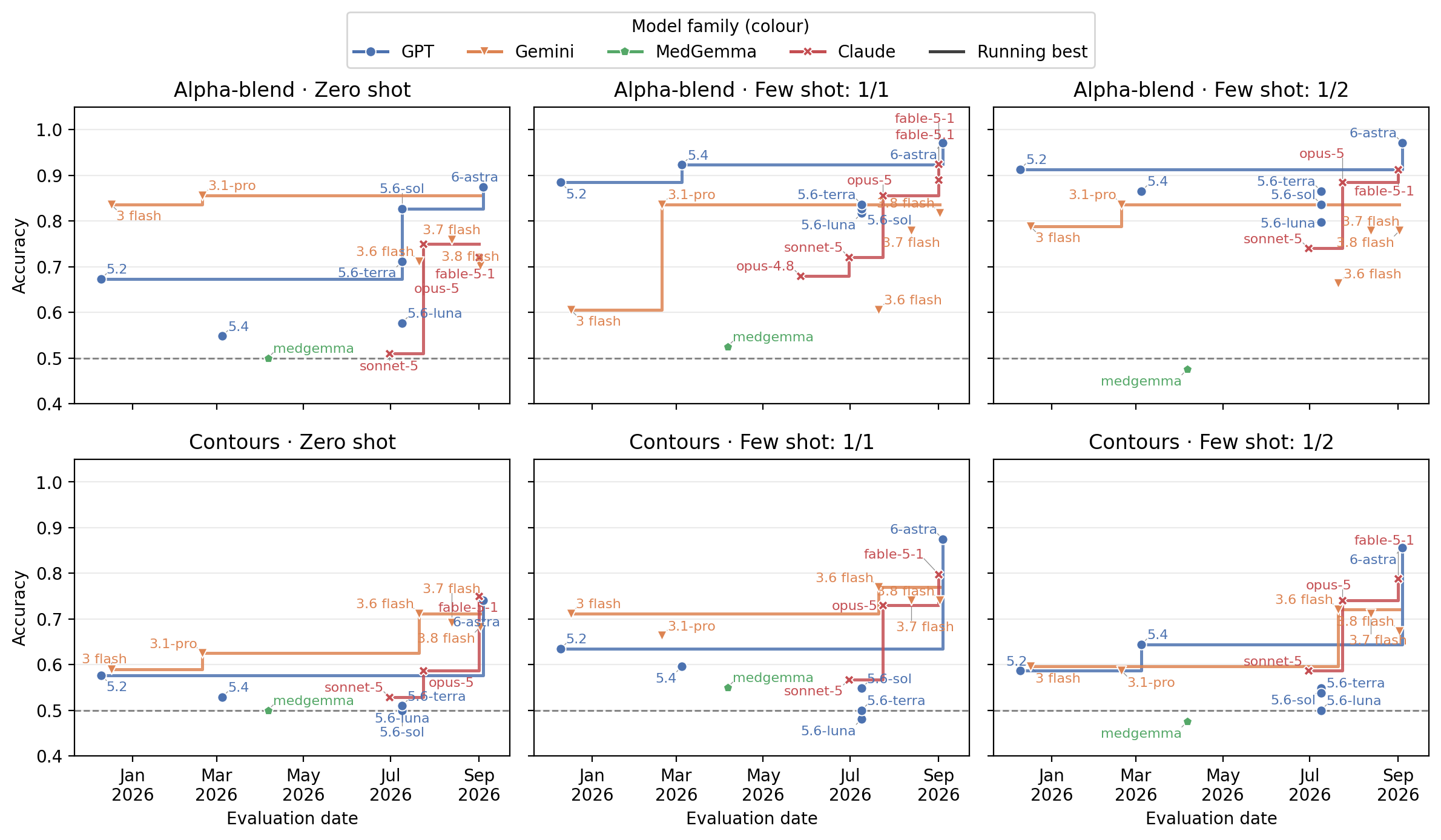} 
    \caption{MLLM misalignment detection performance over time across all six scenarios tested.}
    \label{fig:performance_over_time}
\end{figure}

Broadly speaking, model performance has improved over time, as demonstrated by  \Cref{fig:performance_over_time}. Across most model families and tasks, the best performance was achieved by the most recent frontier models, with the exception of Gemini 3 Flash which shows particularly strong zero-shot performance on the alpha-blend task. 
Local MedGemma performs barely above chance, we also investigated three more open-weight MLLMs that also perform at chance levels (Appendix \ref{sec:app_addtional}).  

\Cref{fig:performance_vs_magnitude_astra,fig:performance_vs_magnitude_luna,fig:performance_vs_magnitude_fable,fig:performance_vs_magnitude_sonnet} show how performance varies with misalignment magnitude for GPT-6 Astra, GPT-5.6 Luna, Claude Fable 5.1 and Claude Sonnet 5. 

% \subsection{Zero vs.\ few shot prompting}
% \label{sec:zero_vs_few_res}
% To explore zero shot vs. few-shot performance, we plot the performance of two models per model family (where relevant), one from the "cost-efficient" set of models (GPT 5.6 Luna, Gemini 3 Flash, Claude Sonnet 5) and one from the "frontier" set of models (GPT 6 Astra, Gemini 3.1 Pro, Claude Fable 5.1). 
% \begin{figure}[ht]
%     \centering
%     \includegraphics[width=0.9\linewidth]{images/model_performance_zero_vs_few.png}
%     \caption{Caption}
%     \label{fig:placeholder}
% \end{figure}

% \subsection{Image presentation method}

\subsection{Experiment 1b: T1w-FLAIR MRI}
\label{sec:exp_1b_res}
Performance is similar across the T1w–FLAIR and MRI–PET experiments as shown in \Cref{tab:results_flair}. Notably, Gemini shows comparatively strong performance in T1w-FLAIR compared to MRI-PET. 
Furthermore, we see some variance in performance across data subsets, though generally models that perform well perform well in all subsets. 

\begin{table}[ht]
    \centering
    \begin{tabular}{c|cccc}
        % \multicolumn{2}{r|}{Image Presentation} & \multicolumn{2}{c}{Alpha-blend}\\
        % \multicolumn{2}{r|}{No.\ Shot Examples (a/m)} & 0/0 & 1/1 \\
         &\multicolumn{3}{c}{No.\ Shot Examples (a/m)}\\
         & \multicolumn{2}{c}{0/0} & \multicolumn{2}{c}{1/1} \\
         Model & T1w-FLAIR & MRI-PET & T1w-FLAIR & MRI-PET \\
         \hline
         GPT-5.6 Luna & 65.0 \footnotesize{(73/59/63)}& \textcolor{gray}{57.7} & 91.3 \footnotesize{(88/86/97)} & \textcolor{gray}{81.7}\\
         GPT-6 Astra & \textbf{91.3} \footnotesize{(81/95/97)} & \textcolor{gray}{87.5} & \textbf{96.3} \footnotesize{(88/100/100)} & \textcolor{gray}{97.1} \\\hline
         Gemini 3.1 Pro & \underline{90.0} \footnotesize{(96/91/84)} & \textcolor{gray}{85.6} & \underline{95.0} \footnotesize{(92/95/97)} & \textcolor{gray}{83.7} \\
         Gemini 3.8 Flash & 86.3 \footnotesize{(88/100/75)} & \textcolor{gray}{70.2} & 91.3 \footnotesize{(88/95/91)} & \textcolor{gray}{81.7} \\\hline
         Claude Sonnet 5 & 53.8 \footnotesize{(62/54/47)} & \textcolor{gray}{51.0} & 66.3 \footnotesize{(69/64/66)} & \textcolor{gray}{72.1} \\
        Claude Fable 5.1 & 76.3 \footnotesize{(73/81/75)} & \textcolor{gray}{72.1} & 86.5 \footnotesize{(81/86/91)} & \textcolor{gray}{89.0} \\\hline
        MedGemma 1.5 4B & 50.0 \footnotesize{(50/50/50)} & \textcolor{gray}{50.0} & 46.3 \footnotesize{(42/50/47)}  & \textcolor{gray}{52.5} \\
    \end{tabular}
    \caption{Misalignment detection in the T1w-FLAIR experiment, compared to equivalent performance on MRI-PET. Small numbers in brackets represent the performance on the A4/ADNI/MICCAI subsets respectively.}
    \label{tab:results_flair}
\end{table}

\subsection{Experiment 1c: Real Misalignment}
\label{sec:exp_1c_res}
Generally speaking, misalignment detection performance in \Cref{sec:exp_1a_res} carries over into misalignment detection of real alignment pipeline failures as shown in \Cref{tab:results_real}, with a notable exception that Gemini 3.8 Flash performs exceptionally well in the zero-shot contours case. 

\begin{table}[ht]
    \centering
    \begin{tabular}{c|cccc}
         \multicolumn{1}{r|}{Image Presentation} & \multicolumn{2}{c}{Contours} & \multicolumn{2}{c}{Alpha-blend} \\
        \multicolumn{1}{r|}{No.\ Shot Examples (a/m)} & 0/0 & 1/1 & 0/0 & 1/1  \\
         \hline
         GPT-5.6 Luna & \result{65.0}{53.3}{100} & \result{60.0}{53.3}{80.0} & \result{68.0}{60.0}{90.0} & \result{85.0}{97.0}{50.0} \\
         GPT-6 Astra & \result{80.0}{73.0}{100} & \result{80.0}{73.0}{100} & \result{85.0}{80.0}{100} & \result{90.0}{87.0}{100} \\\hline
         Gemini 3.1 Pro & \result{80.0}{90.0}{50.0} & \result{80.0}{100}{20.0} & \result{80.0}{90.0}{50.0} & \result{85.0}{97.0}{50.0} \\
         Gemini 3.8 Flash & \result{93.0}{90.0}{100} & \result{80.0}{77.0}{90.0} & \result{88.0}{100}{50.0} & \result{85.0}{100}{40.0} \\\hline
        Claude Fable 5.1 & \result{85.0}{90.0}{70.0} & \result{80.0}{80.0}{80.0} & \result{75.0}{70.0}{90.0} & \result{85.0}{83.0}{90.0} \\
        Claude Sonnet 5 & \result{68.0}{57.0}{100} & \result{83.0}{80.0}{90.0} & \result{78.0}{77.0}{80.0} & \result{75.0}{80.0}{60.0} \\\hline
        
        MedGemma 1.5 4B & \result{25.0}{0.0}{100} & \result{75.0}{100}{0} & \result{25.0}{0.0}{100}  & \result{50.0}{81.3}{70.0} \\
    \end{tabular}
    \caption{Misalignment detection in the real cases. Values reported are accuracy as a percentage with sensitivity/specificity below. MedGemma shows always predicting MISALIGNED achieves 75\% accuracy. }
    \label{tab:results_real}
\end{table}

\subsection{Experiment 2: Comparison to local models}
\label{sec:exp2_results}
\begin{figure}
    \centering
    \includegraphics[width=0.65\linewidth]{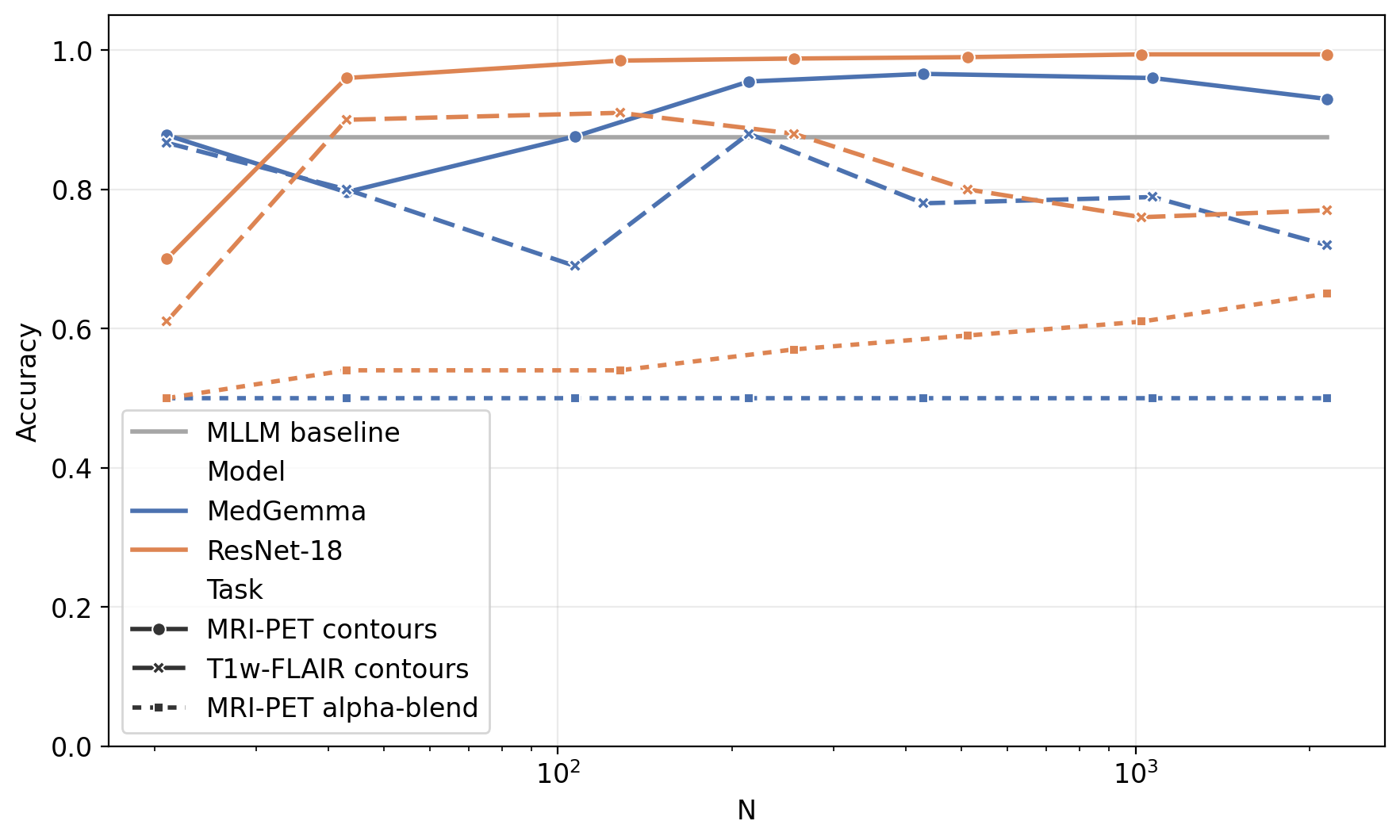}
    \caption{Local model performance with training set size for both the within-distribution test set (MRI-PET) and out-of-distribution test set (T1w-FLAIR). Grey line represents best frontier MLLM performance on the same task (MRI-PET with contour image presentation)}
    \label{fig:local_models}
\end{figure}

Both MedGemma and ResNet-18 can achieve performance on-par with or exceeding frontier MLLMs on the task they are tuned for (MRI-PET) as shown in \Cref{fig:local_models}. 
Both models show reasonable generalisability to the new modality (T1w-FLAIR) with a similar trend that as dataset size increases, generalisability falls. 
However, neither model generalises well to the new image-presentation mode (MRI-PET alpha-blend). 

\section{Discussion}
In this work, we have demonstrated that frontier MLLMs can be effective medical image alignment assessors. 
The most recent frontier models, Claude Fable 5.1 and GPT-6 Astra show strong performance in almost all settings. This is notably true of GPT-6 Astra which performs best in 7 of the 8 settings and second best in the remaining one. 

\paragraph{Model performance over time} As seen in \Cref{fig:performance_over_time}, MLLM misalignment assessment has generally improved over time as more and more capable models are released. In the alpha-blend image presentation case, model performance has been strong even with relatively old models such as Gemini 3 Flash and GPT-5.2, however it is only recently that frontier MLLMs can perform well in the contours setting. 

\paragraph{Zero-shot vs. few-shot} There is a general trend of few-shot examples helping frontier MLLM performance, particularly when using alpha-blend image presentation as seen in \Cref{tab:results}. 
Almost universally, showing the model one aligned and one misaligned example brings a performance increase over the zero-shot case, however adding an additional misaligned example brings marginal gains.
Interestingly, showing examples is not always guaranteed to improve performance; Gemini 3 Flash presents an unusual case where zero-shot performance in the alpha-blend case is strong, however performance degrades with examples. 

\paragraph{Image presentation} For almost all models, alpha-blend image presentation enables more accurate detection of misalignment compared to contours. 
In particular, only the most recent frontier MLLMs achieve strong performance on contours even with examples, while most models achieve strong performance on alpha-blend.
One reason for this may that alpha-blend presentation makes edge misalignment apparent, while contours may be more useful in subtle internal brain misalignment as demonstrated by the examples in \Cref{fig:idea_example}.

\paragraph{Model performance vs. misalignment magnitude}
As shown in \Cref{fig:performance_vs_magnitude_astra,fig:performance_vs_magnitude_luna,fig:performance_vs_magnitude_fable,fig:performance_vs_magnitude_sonnet}, model detection performance varies with misalignment magnitude, with only the top performing models able to detec small misalignment reliably. For instance, \Cref{fig:performance_vs_magnitude_astra} shows that although GPT-6 Astra achieves strong overall accuracy in zero-shot contours, it is only reliably detecting large misalignments. 

\paragraph{Real misalignment detection} Recent frontier models that perform strongly in \Cref{sec:exp_1a_res} also reliably detect real misalignment, demonstrating that this synthetic experiment represents real misalignment well. Consistent with Experiment 1A, GPT-6 Astra acheives a good balance of sensitivity and specificity, while Gemini 3.8 Flash performs particularly strongly on the zero-shot contours in this case.

\paragraph{T1w-FLAIR assessment} MLLMs generally perform slightly better on the T1w-FLAIR alignment task than the MRI-PET task, with some exceptions such as GPT-6 Astra and Claude Fable 5.1 in few-shot mode where they already perform very strongly on the MRI-PET task. 
This result is not unexpected, since T1w and FLAIR are both MRI modalities that can reveal different aspects of anatomy, while PET measures other physiological and molecular information so the assessment of alignment with MRI is expected to be more inherently difficult. 

We see some amount of variance across the data subsets (A4/ADNI/MICCAI) as seen in \Cref{tab:results}. 
One thing to note is that the single example subject given in the few shot case comes from the MICCAI challenge subset and we see in almost all models that the MICCAI subset is the best performing one. 
This suggests that although few-shot prompting can generally help MLLMs, we need to carefully consider which examples we select.
Indeed for the Gemini models, while overall model performance improves with examples, performance in some subsets (A4 for 3.1 Pro, ADNI for 3 Flash) actually falls when given examples from a different dataset.

\paragraph{Comparison to local models}
Both local models tested achieve performance at or above the level of frontier MLLMs on the MRI-PET contour task at almost all training dataset sizes, with the exception of 1\% (21 subjects) for ResNet-18 and 2\% (43 subjects) for MedGemma. 
Eventually, both models achieve near perfect performance with accuracies \textgreater98\%. The best frontier MLLM only achieves 85.6\% on this task, with examples, demonstrating that task-specific models can outperform generalist ones. 

Both models show reasonable performance on the T1w-FLAIR contours task across most training dataset sizes. 
This relatively strong performance on a modality for which they were not trained suggests that they are learning features that can detect misalignment of contours on top of multiple base image types. 
The T1w-FLAIR performance falls off as training dataset size grows in both cases, however, suggesting that the models are potentially learning more task-specific features at the cost of reduced generalisability when seeing more data. 

Neither model generalises well to different image presentation modes when tested on the MRI-PET alpha-blend task. 
MedGemma performs no better than chance, always predicting "aligned" while ResNet-18 shows an interesting, inverse behaviour to T1w-FLAIR contour generalisability.
As the training dataset size grows, MRI-PET alpha-blend performance grows while T1w-FLAIR contours performance drops. 

One important consideration when weighing up API-based frontier models and local models is that although there may be generalisability concerns with local models, API-based models require sending data to third parties. 
This is of particular relevance for medical imaging data, where patient confidentiality may prevent the use of API-based models. 
Furthermore, the frontier models that perform strongest are also the most expensive per-token, while local models are cheaper to run after the upfront cost of hardware. 

\paragraph{Limitations and Future Work}
Currently, this work only investigates two exemplar image alignment tasks chosen to be representative of the myriad use cases of image alignment assessment in medical imaging. 
Future work will extend this approach to other applications including within-subject motion detection and between-subject alignment for group-level analyses. 

Furthermore, for practicality and cost reasons while piloting this approach across a wide range of methods, dataset sizes were kept reasonably small. 
The results of this work, however, will enable us to narrow down model choices moving forward and analyse more data. 

A limitation of the approach taken in this study is that the misalignment we are assessing is synthetically introduced through random rigid transformations. 
This may not realistically reflect the kinds of misalignments we see in practice when using automated alignment techniques. 
For this study, since we are interested in methods that use rigid transformations to achieve alignment, we believe that the synthetic misalignment approach is reasonable and will result in realistic samples. 
A more careful approach will be required when considering other alignment assessment tasks such as those involving non-linear deformations.

\section{Conclusion}
\label{sec:conclusion}
We have demonstrated that medical image alignment assessment can be a useful test bed for generalist visual reasoning. 
Performance varies substantially across model generations, with only the most recent frontier models performing strongly across most scenarios. 
However, contour based image presentation remains a challenge, with a maximum of only 75\% accuracy in the zero-shot setting. 

The comparison with locally finetuned models shows a trade-off. 
They are able to match or outperform frontier MLLMs on the task that they are tuned for; however, this comes at the expense of reduced generalisability onto different tasks.
Although local modals have an advantage in keeping data local, they need to be carefully tuned and validated for each use case. 

These results support the use of MLLMs as part of medical imaging pipelines, providing evidence that these models can perform effective visual alignment assessments. 
Evaluation across more modalities and alignment tasks will be important next steps to establish MLLMs as routine visual image assessors. 
For now though, the strong performance of frontier MLLMs on these exemplar tasks is evidence of the promise of these methods for visual medical image assessment. 

% \subsection*{AI use statement}
% In this work, we used generative AI tools for retrieval and discovery of related work, suggestion of modifications to figures, creation and editing of software code and suggestion of changes to text to improve readability.
% We have not used generative AI tools to generate synthetic data sets, propose or refine hypotheses, design or provide feedback on research  methodology or experiments, implement methods, clean and reformat dataset, support qualitative and thematic data analysis, interpret results,
% and help develop theoretical models or conceptual frameworks, formulate mathematical claims, provide critical ingredients for proving mathematical claims, assist in the writing of proofs, assist with translation, are not applicable to this work.
% We have reviewed all AI-assisted work, checking AI proposed changes to text and figures before implement and reviewing AI generated code. 
% We take responsibility for the final content of this work,
% including text, claims or artifacts produced with the aid of generative AI.

% \subsection*{Reproducibility statement}
% We aim to make our work reproducible by providing detailed descriptions of our methods, data processing and experimental design. 
% We provide examples of the prompts used in this work in \Cref{sec:appendix} to aid in reproducibility. 

\subsubsection*{Author Contributions}
\paragraph{Ross Callaghan:} Conceptualisation, Methodology, Software, Investigation, Visualisation, Writing – Original Draft, Writing - Review \& Editing

\paragraph{Niannu Gao:} Software, Investigation

\paragraph{Hojjat Azadbakht:} Writing – Review \& Editing, Supervision, Resources, Funding Acquisition

\paragraph{Hui Zhang:} Conceptualisation, Methodology, Writing – Review \& Editing, Supervision, Resources, Funding Acquisition.

\subsubsection*{Acknowledgments}
This work is supported by the National Institute for Health and Care Research Grant NIHR207476. 

Data collection and sharing for the Alzheimer's Disease Neuroimaging Initiative (ADNI) is
funded by the National Institute on Aging (National Institutes of Health Grant
U19AG024904). The grantee organization is the Northern California Institute for Research
and Education. In the past, ADNI has also received funding from the National Institute of
Biomedical Imaging and Bioengineering, the Canadian Institutes of Health Research, and
private sector contributions through the Foundation for the National Institutes of Health
(FNIH) including generous contributions from the following: AbbVie, Alzheimer’s Association;
Alzheimer’s Drug Discovery Foundation; Araclon Biotech; BioClinica, Inc.; Biogen; BristolMyers Squibb Company; CereSpir, Inc.; Cogstate; Eisai Inc.; Elan Pharmaceuticals, Inc.; Eli
Lilly and Company; EuroImmun; F. Hoffmann-La Roche Ltd and its affiliated company
Genentech, Inc.; Fujirebio; GE Healthcare; IXICO Ltd.; Janssen Alzheimer Immunotherapy
Research \& Development, LLC.; Johnson \& Johnson Pharmaceutical Research \&
Development LLC.; Lumosity; Lundbeck; Merck \& Co., Inc.; Meso Scale Diagnostics, LLC.;
NeuroRx Research; Neurotrack Technologies; Novartis Pharmaceuticals Corporation; Pfizer
Inc.; Piramal Imaging; Servier; Takeda Pharmaceutical Company; and Transition
Therapeutics.

The A4 Study was a secondary prevention trial in preclinical Alzheimer's disease, aiming to slow cognitive decline associated with brain amyloid accumulation in clinically normal older individuals. The A4 Study was funded by a public-private-philanthropic partnership, including funding from the National Institutes of Health-National Institute on Aging, Eli Lilly and Company, Alzheimer's Association, Accelerating Medicines Partnership, GHR Foundation, an anonymous foundation, and additional private donors, with in-kind support from Avid Radiopharmaceuticals, Cogstate, Albert Einstein College of Medicine and the Foundation for Neurologic Diseases.The companion observational Longitudinal Evaluation of Amyloid Risk and Neurodegeneration (LEARN) Study was funded by the Alzheimer's Association and GHR Foundation. The A4 and LEARN Studies were led by Dr. Reisa Sperling at Brigham and Women's Hospital, Harvard Medical School, and Dr. Paul Aisen at the Alzheimer's Therapeutic Research Institute (ATRI) at the University of Southern California. The A4 and LEARN Studies were coordinated by ATRI at the University of Southern California, and the data are made available under the auspices of Alzheimer’s Clinical Trial Consortium through the Global Research \& Imaging Platform (GRIP). The complete A4 Study Team list is available on: https://www.actcinfo.org/a4-study-team-lists/. We would like to acknowledge the dedication of the study participants and their study partners who made the A4 and LEARN Studies possible.

\bibliography{sample}
\bibliographystyle{iclr2027_conference}

\appendix
\section{Appendix}
\counterwithin{figure}{section}
\renewcommand{\thefigure}{\thesection\arabic{figure}}
\label{sec:appendix}
\subsection{Example Prompts}
\label{sec:appendix_prompts}
\begin{promptbox}{Zero-shot contours prompt}
This image contains three views of the same brain: sagittal, coronal, and axial.

The grayscale background is a PET image. The coloured anatomical contours come from a T1-weighted MRI segmentation.

Determine whether the PET image and the T1-weighted MRI are spatially aligned.

Return exactly one label:
ALIGNED
MISALIGNED

Do not explain your answer.

<test_image>
\end{promptbox}

\begin{promptbox}{Zero-shot alpha-blend prompt}
The image contains three views of the same brain: sagittal, coronal, and axial.

Each view is an alpha blend: PET is the base image and the coloured T1-weighted
MRI is overlaid at 0.5 opacity.

Determine whether the PET image and the T1-weighted MRI are spatially aligned.

Return exactly one label:
ALIGNED
MISALIGNED

Do not explain your answer.

<test_image>
\end{promptbox}

\begin{promptbox}{Few-shot prompt template}
<zero-shot template> 

Example 0 (known classification):
<example_0_img>
Known answer for example 0: <label_0>

Example 1 (known classification):
<example_1>
Known answer for example 1: <label_1>

Test case: classify only the following image(s), using the examples above as guidance.
<test_image>
Answer for the test case:
\end{promptbox}

\newpage
\section{Additional Open-Weight Model Performance}
\label{sec:app_addtional}
\begin{table}[!htb]
    \centering
    \begin{tabular}{c|cccc}
    \multicolumn{1}{r|}{Image Presentation} & \multicolumn{2}{c}{Contours} & \multicolumn{2}{c}{Alpha-blend} \\
    \multicolumn{1}{r|}{No.\ Shot Examples (a/m)} & 0/0 & 1/1 & 0/0 & 1/1 \\
    Model & & & &\\
    \hline
    InternVL3 2B & \result{50.0}{0.0}{100} & \result{50.0}{100.0}{0.0} & \result{50.0}{0.0}{100} & \result{50.0}{100}{0.0} \\
    Qwen3-VL 8B & \result{50.0}{100}{0.0} & \result{50.0}{100.0}{0.0} & \result{50.0}{100}{0.0} & \result{50.0}{100}{0.0} \\
    LLaVA-1.5 7B & \result{50.0}{0.0}{100} & \result{50.0}{0.0}{100} & \result{50.0}{0.0}{100} & \result{50.0}{0.0}{100} \\
    \hline

\end{tabular}

    \caption{Misalignment detection accuracy across additional open-weight MLLM models, zero/few shot prompting and image presentation method. a/m refers to number of aligned/misaligned examples used in the few-shot case. Values reported are accuracy as a percentage with sensitivity/specificity below.}
    \label{tab:results_additional}
\end{table}

\section{Model Performance vs. Misalignment Magnitude}
\begin{figure}[htb]
    \centering
    \includegraphics[width=0.8\linewidth]{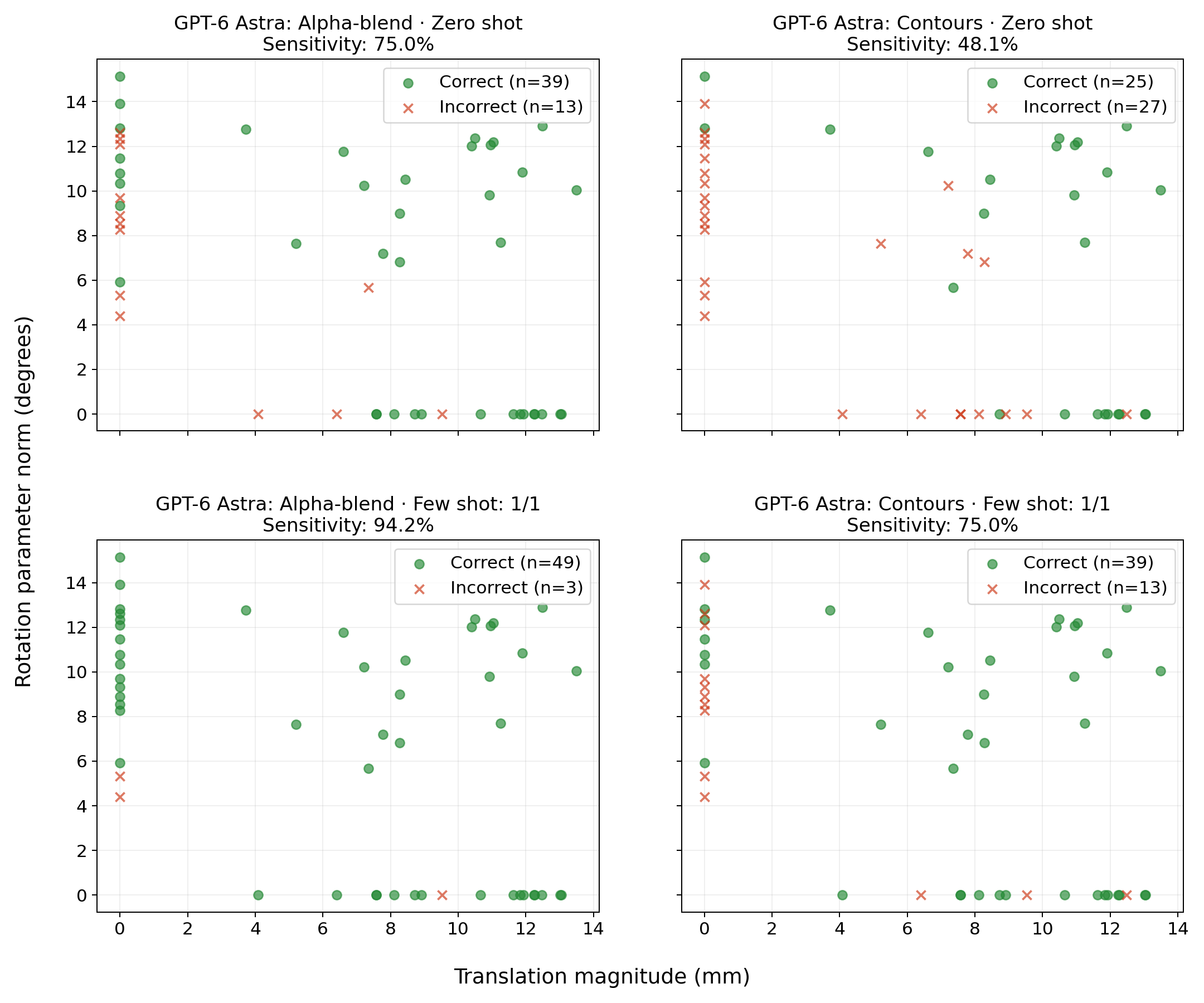}
    \caption{GPT-6 Astra misalignment detection performance against misalignment magnitude. }
    \label{fig:performance_vs_magnitude_astra}
\end{figure}

\begin{figure}[htb]
    \centering
    \includegraphics[width=0.8\linewidth]{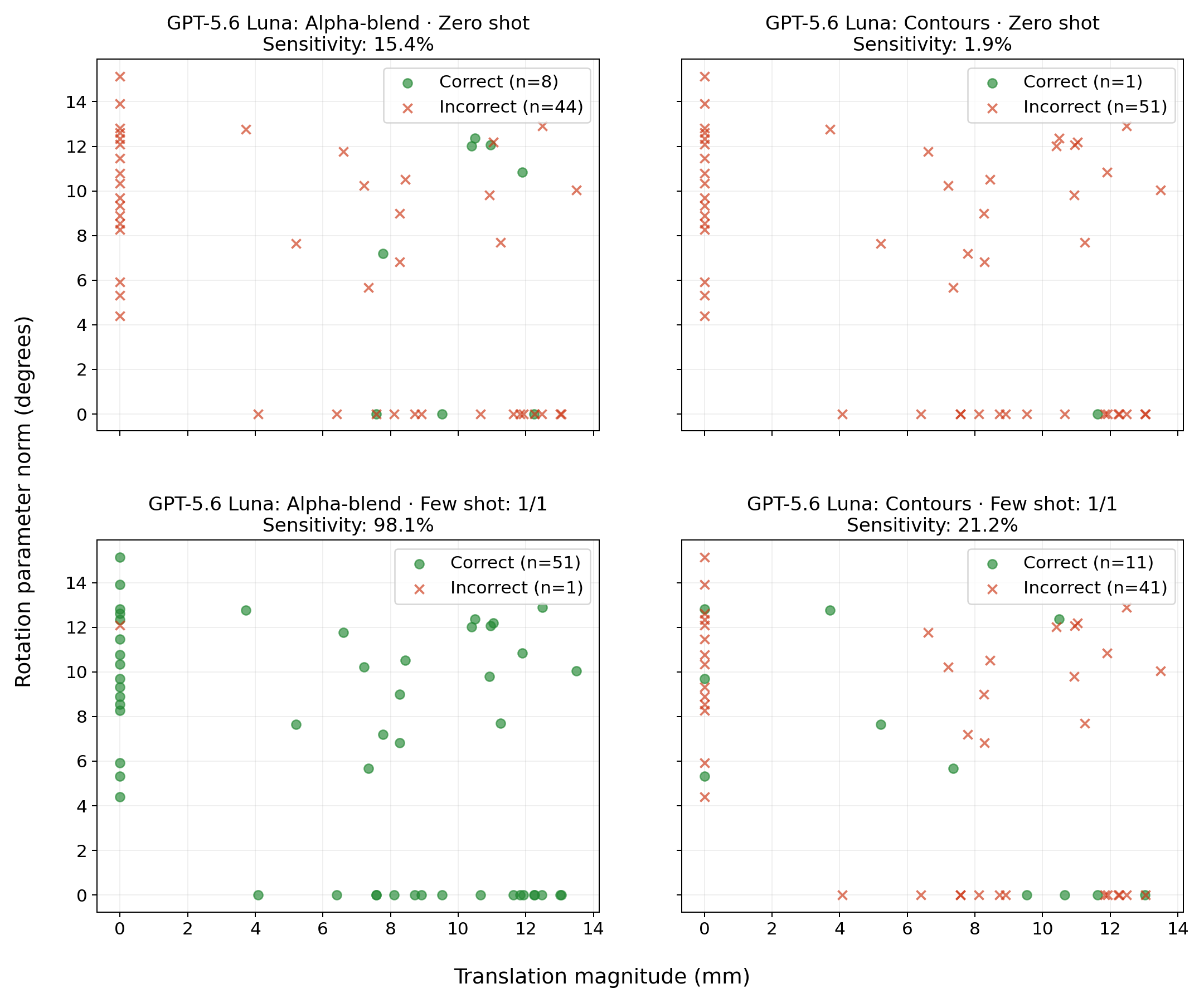}
    \caption{GPT-5.6 Luna misalignment detection performance against misalignment magnitude. }
    \label{fig:performance_vs_magnitude_luna}
\end{figure}

\begin{figure}[htb]
    \centering
    \includegraphics[width=0.8\linewidth]{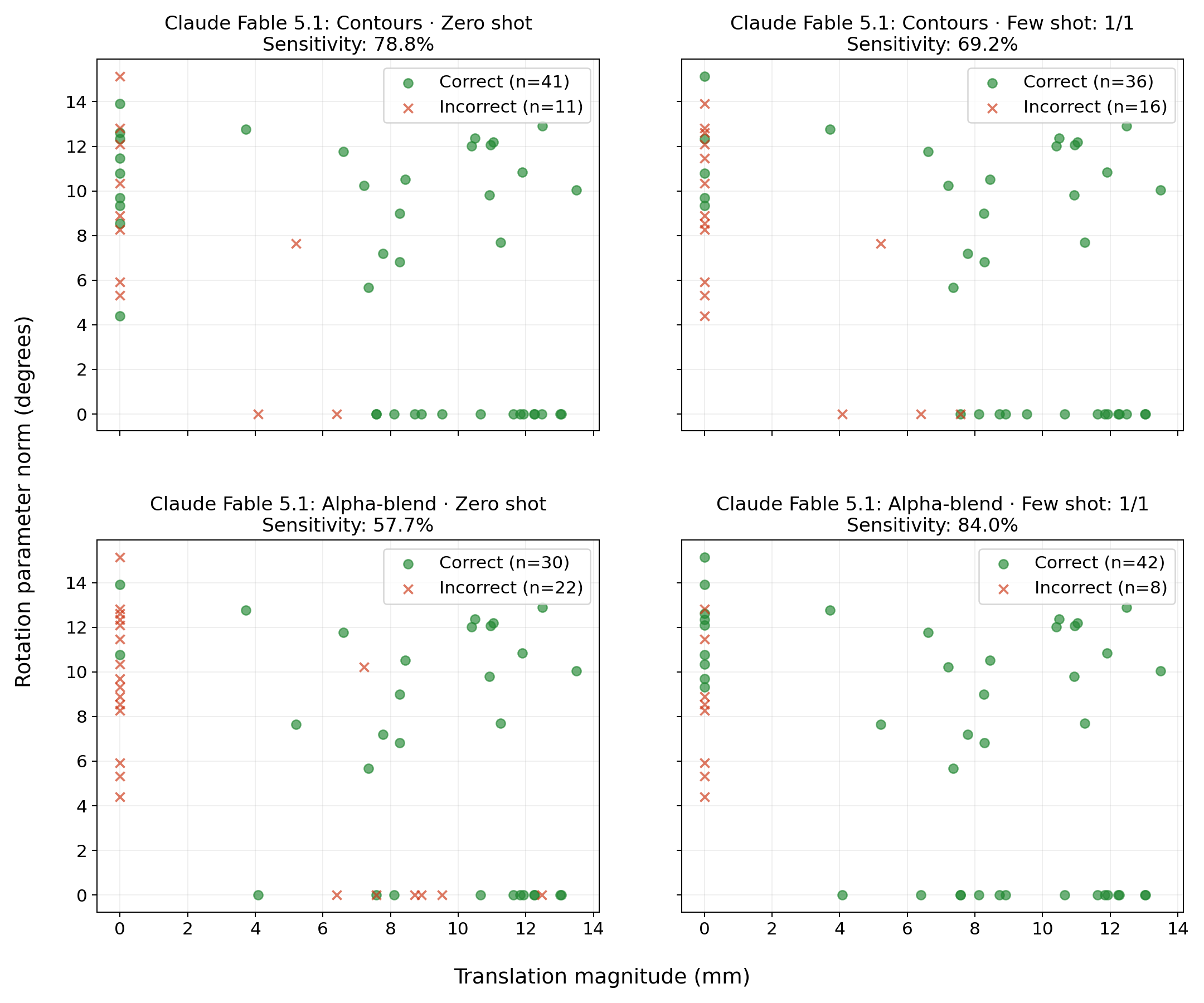}
    \caption{Claude Fable 5.1 misalignment detection performance against misalignment magnitude. }
    \label{fig:performance_vs_magnitude_fable}
\end{figure}

\begin{figure}[htb]
    \centering
    \includegraphics[width=0.8\linewidth]{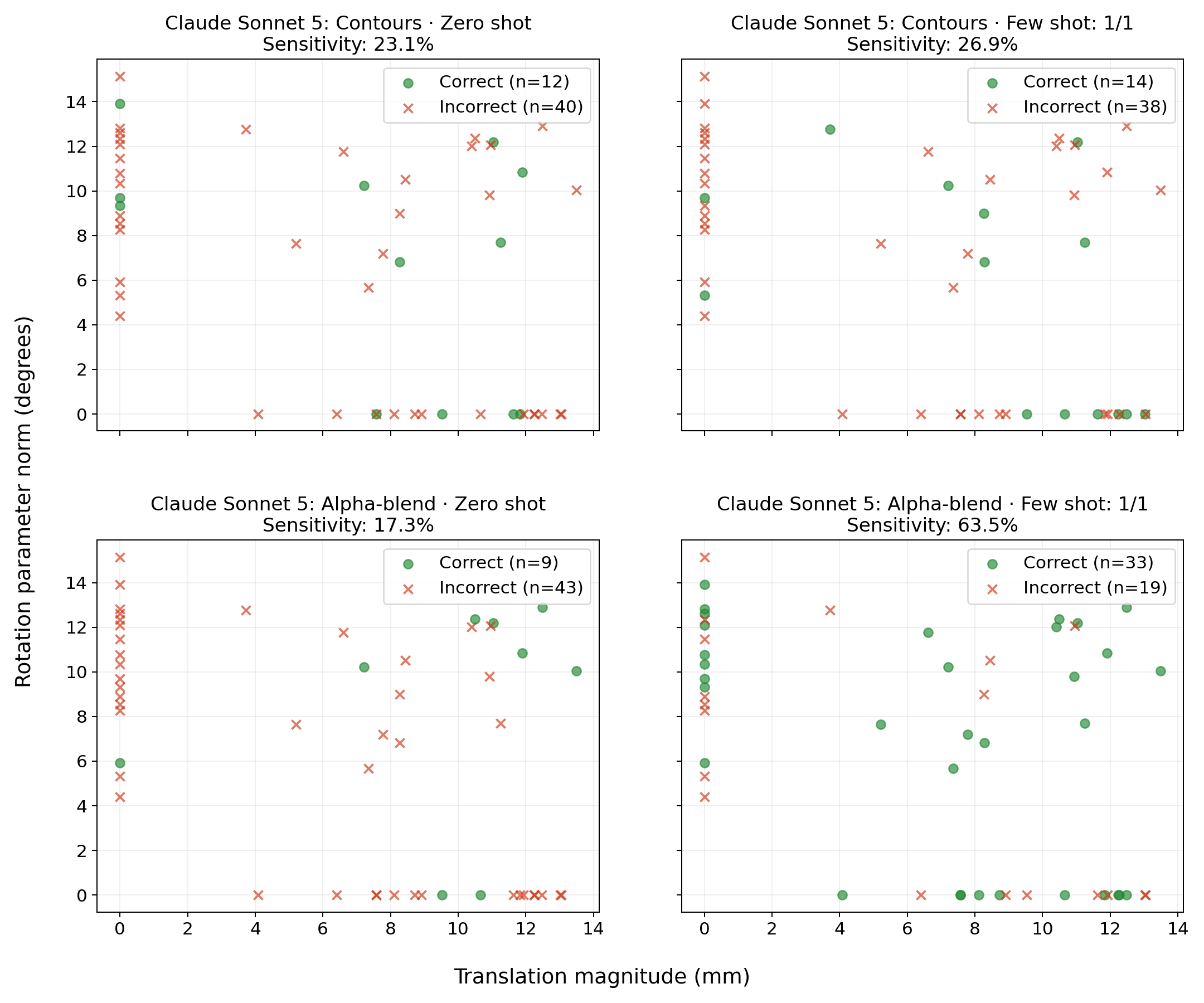}
    \caption{Claude Sonnet 5 misalignment detection performance against misalignment magnitude. }
    \label{fig:performance_vs_magnitude_sonnet}
\end{figure}

\end{document}